\pdfoutput=1
\documentclass[11pt]{article}

\usepackage[]{naacl2021}

\usepackage{times}
\usepackage{latexsym}
\usepackage{graphicx}

\usepackage{multirow}
\usepackage{bbm}
\usepackage{amsmath}
\usepackage{svg}

\usepackage{listings}
\usepackage{xcolor}

\usepackage{subfig}

\usepackage{listings}

\def\shortname {ToolWriter }

\usepackage[T1]{fontenc}
\usepackage[utf8]{inputenc}

\usepackage{microtype}

\title{Generate, Transform, Answer: Question Specific Tool Synthesis for Tabular Data}
\author{Carlos Gemmell \\
  University of Glasgow \\
  {\tt c.gemmell.1@research.gla.ac.uk  } \\\And
  Jeffrey Dalton \\
  University of Glasgow \\
  { \tt  jeff.dalton@glasgow.ac.uk} \\
}

\begin{document}
\maketitle
\begin{abstract}

Tabular question answering (TQA) presents a challenging setting for neural systems by requiring joint reasoning of natural language with large amounts of semi-structured data. Unlike humans who use programmatic tools like filters to transform data before processing, language models in TQA process tables directly, resulting in information loss as table size increases. In this paper we propose \shortname to generate query specific programs and detect when to apply them to transform tables and align them with the TQA model's capabilities. Focusing \shortname to generate row-filtering tools improves the state-of-the-art for WikiTableQuestions and WikiSQL with the most performance gained on long tables. By investigating headroom, our work highlights the broader potential for programmatic tools combined with neural components to manipulate large amounts of structured data.

\end{abstract}

\section{Introduction}
An important area for research in large language models (T5, PaLM, GPT-3) is combining them with "tools" to enhance their capabilities in question answering\cite{Schick2023ToolformerLM, gao2022palprogramaided, parisi2022talmtoolaugmented, Lazaridou2022InternetaugmentedLM}. Tool-augmented approaches enable language models to externalize knowledge and computation by making explicit calls to APIs. However, these approaches do not process semi-structured data. We show that current models degrade in effectiveness significantly when questions and  data become long and complex. A key task that demonstrates these limitations is tabular question answering (TQA) where long tables and complex questions are particularly challenging for current models. 

% -- related and alternatives   
% Backgound and motivation --> Introduce tool use for QA -- PAL, Toolformer, TALM --> what's the gap here vs those models.   TQA here and why it's important and the right task setup.  Summary of how thse models fail on TQA. 

\begin{figure}
\centering
  \includegraphics[width=8cm]{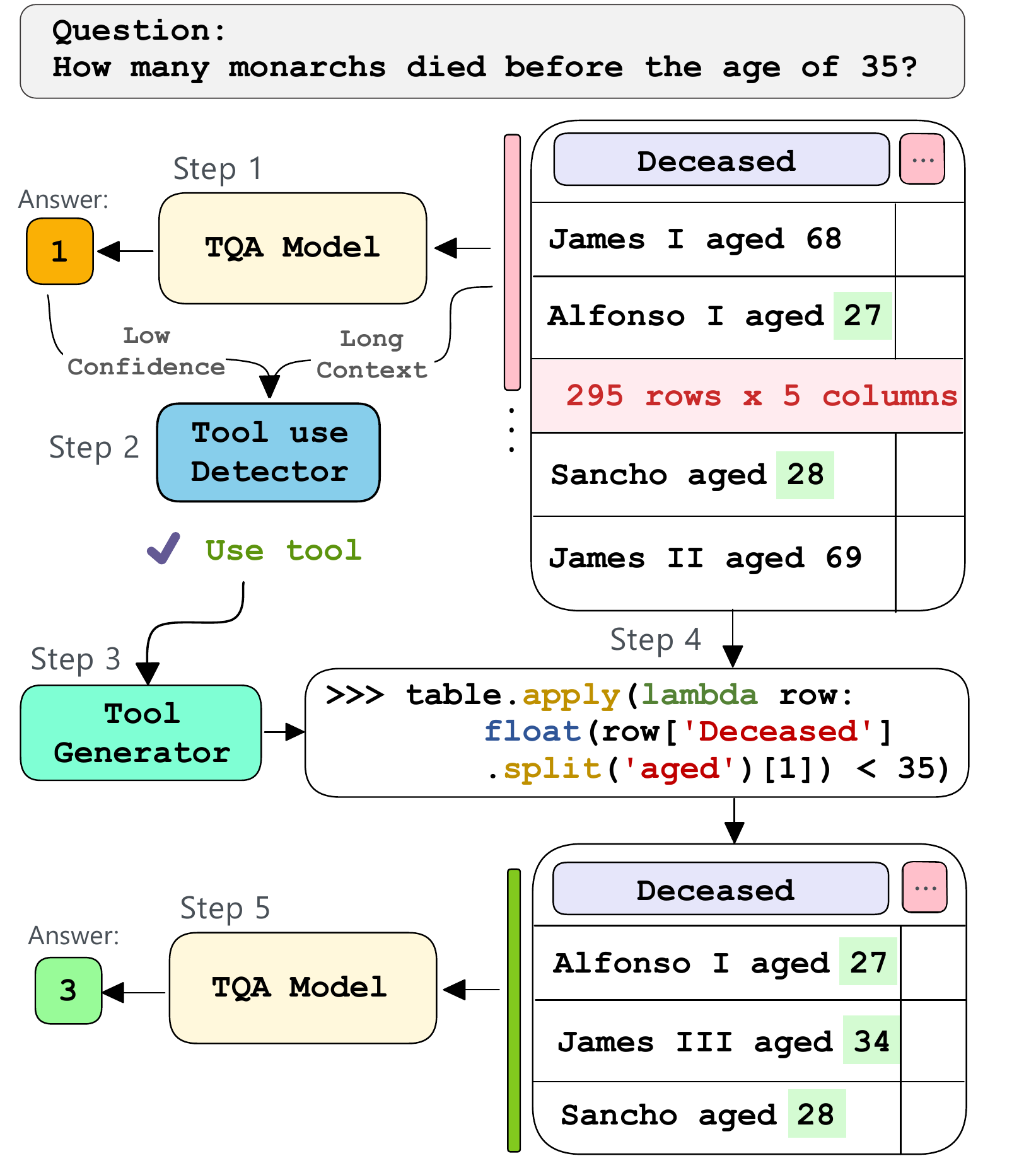}%
  \caption{\shortname for tabular question answering introduces: 1) A tool-use detector; 2) A tool generator. Here a row-filter tool is generated as a program that transforms semi-structured data.}
  \label{main_figure}
\end{figure}

Tabular question answering is a task in natural language processing that involves leveraging information from a semi-structured table to answer multi-hop compositional questions.
It discourages purely symbolic approaches due to latent structure captured in natural language in the table. As shown in Figure \ref{main_figure} strings in cells often contain numerical values implicitly contextualized by surrounding text like "aged".

Current TQA systems \cite{Xie2022UnifiedSKGUA, Jiang2022OmniTabPW, Liu2021TAPEXTP} linearize tables as a token string and process it jointly with the question. In response to long tables present in WikiTableQuestions \cite{Pasupat2015CompositionalSP} and WikiSQL \cite{zhong2017seq2sqlgenerating} language models for TQA increase the context size to 1024 tokens incurring a high memory cost. However, we show in Section \ref{analysis} as table size increases average model performance significantly degrades by 40\%. Moreover, 23\% of tables in WikiTableQuestions are truncated at 1024 tokens.

We propose \shortname, a new method that augments language model capabilities. In response to a question over a table, \shortname decides if tool use is required. It then generates a program to transform the table to simplify it to make question answering more effective. The generated tools are code that can be applied to all kinds of tables regardless of size which overcomes key language model limitations. The proposed ToolWriter approach is model agnostic and can be flexibly combined with existing models in a zero-shot setup.

% Experimental setup and results
% -- Summarize key findings here.
In this work, we compare multiple detection strategies and tool generation approaches to generate query and table-specific Python programs. \shortname leverages our best combination for tool generation (zero-shot GPT-3) and tool-use detection (combined answer confidence and table length). Our method improves the state-of-the-art exact match results on WikiTableQuestions to 64.9 (+1.9\%) and WikiSQL to 90.5 (1.5\%).

We summarize our contributions as the following:
\begin{itemize}
    \item We characterize the behavior of language models for TQA on WikiTableQuestions and find significant performance degradation as table size increases.

    \item We propose \shortname to detect when tool-use is required and generate query and table-specific programs as tools to transform tabular data. By generating row-filter tools we achieve new state-of-the-art results on WikiTableQuestions and WikiSQL.

    \item Through ablation studies analysis we show all our tool generation methods are model agnostic and improve effectiveness as table size increases.
\end{itemize}

\section{Task Definition}
\label{tasd_def}

Tabular question answering is a task in natural language processing that involves leveraging information from a semi-structured table $T$ to generate an answer $\hat{y}$ to a question $q$. Questions are expressed in natural language and implicitly involve compositional types of reasoning to access and aggregate information in the table. These questions are implicitly multi-step and require a combination of symbolic reasoning and natural language understanding.

A system has access to a training set $D = \{(x=(q, T), \hat{y})\}$ of questions, tables, and answers. Tables between training and evaluation are disjoint to prevent memorization. The only restriction on the question is that it must be answerable given the information provided in the table. Average exact match (EM) over D between the predicted answer $\hat{y}$ and target $y$ is used as the primary metric.

\subsection{Datasets}

% \red{We use standard datasets and use XYZ models...}

\textbf{WikiTableQuestions} \cite{pasupat2015compositionalsemantic} serves as our initial exploration into the limitations of current models. It is a tabular question-answering dataset from 2,108 HTML tables and crowdsourced question-answer pairs. Despite multiple questions per table in both train and test settings, tables between the training and testing set are distinct. WikiTableQuestions boasts several key attributes that make it an effective and challenging benchmark:

\begin{itemize}
    \item Questions often require multiple steps to answer by gathering distinct pieces of information from a single table.
    \item Tables are not perfectly formatted often displaying non-consistent cell values depending on the implicit capabilities of the reader to discern different sections. 
    \item Cells often contain interleaved formal representations and natural language making use of pure programmatic approaches challenging. 
\end{itemize}

\textbf{WikiTableQuestions-Filter} is a subset of the WikiTableQuestions dev set for analysis in Section \ref{simple_queries_results} to isolate tool performance in \shortname independent from the detector. Leveraging SQUALL annotations \cite{shi2020onthepotential} we keep samples that contain a \texttt{WHERE} clause after a \texttt{SELECT}. This results in 1256 question-table-answer triplets.

\textbf{WikiSQL} \cite{zhong2017seq2sqlgenerating}, similar to WikiTableQuestions, consists of 80,654 question-answer pairs over 24,241 tables from Wikipedia. Although its original intention was for semantic parsing it has been adapted to weak supervision settings by just using the target answer span as the source of signal. All tables in the dataset are fully parseable with types. Questions are simpler compared to WikiTableQuestions and only contain operations on full cell values that are fully parseable by an SQL query.

\subsection{Baseline models}

We investigate the limitations systems with varying task specific supervision. 

\textbf{BART} \cite{lewis2020bartdenoisingsequence} is a Transformer \cite{vaswani2017attention} pre-trained with a denoising objective. For TQA it is fine-tuned with 1024 tokens of context jointly processing the query and a linearized table as follows: $x = q \texttt{[HEAD]}, c1, · · ·, cN , \texttt{[ROW]}, 1, r1, \texttt{[ROW]}, 2, r2.$.

\textbf{TapEx} \cite{liu2021tapextablepre} is a BART model fine-tuned to mimic an SQL executor on 5 million grammar-generated SQL statements. TapEx is currently state-of-the-art on the weak formulation of WikiSQL.

\textbf{Omnitab} \cite{Jiang2022OmniTabPW} is based on TapEx and further fine-tuned on natural language. The pre-training translates synthetic SQL queries into questions and mines similar passages to tables for masked language modeling. OmniTab is state-of-the-art on WikiTableQuestions and may be seen as the narrowest model for TQA due to its fine-tuning regime.

\textbf{UnifiedSKG} \citet{xie2022unifiedskgunifying} is a T5 transformer \cite{raffel2020exploringthelimits} with a standardized multi-task text-to-text format on structured knowledge (tables, knowledge bases, semantic parsing, etc...). 

\textbf{FlanT5} \cite{wei2021finetunedlanguage} takes a middle-ground approach between strong supervision and generality by instruction-tuning transformers on 62 different types of NLP tasks. Through in-context learning, it provides a strong baseline for TQA.

\textbf{GPT-3} \cite{Brown2020LanguageMA} showcases an unsupervised in-context learning approach to TQA. GPT-3 shows strong performance in TQA with zero-shot Chain-of-Thought (CoT) reasoning explicitly answering step-by-step \cite{kojima2022largelanguagemodels, wei2022chainofthought, Chen2022LargeLM}.

\section{Behaviour Analysis}
\label{analysis}

Tabular question answering is a challenging setting since tables can be exceedingly long. Context length in Transformer architectures is often limited to 512 tokens due to a quadratic memory cost \cite{vaswani2017attention}. In WikiTableQuestions 41.7\% of linearised tables exceed 512 tokens without even considering the question tokens. Current approaches to TQA patch this problem by increasing the context limit to 1024 tokens (Omnitab, TapEx, UnifiedSKG) and 2048 in GPT-3.  This incurs a significant memory cost often prohibiting the use of such models. However, even at 1024 tokens of context 23.8\% of tables are truncated thus incurring data loss. 

\subsection{Effectiveness across table size}
\begin{figure}
\includegraphics[width=0.5\textwidth]{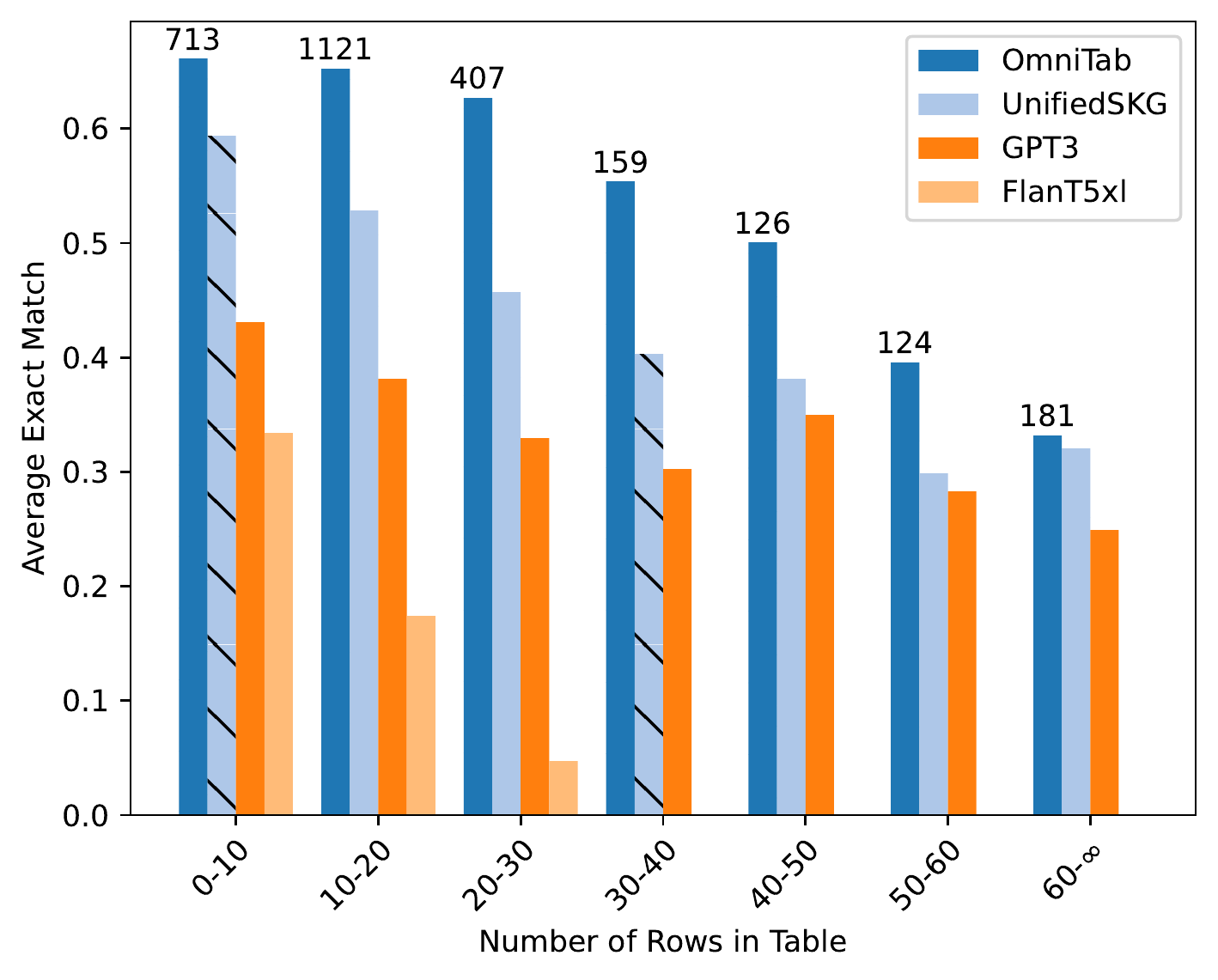}
\caption{Exact match by table size on the WikiTableQuestions dev set. The number of dataset samples per row subset is shown above each bar.}
\label{size_performance_analysis}
\end{figure}

Figure \ref{size_performance_analysis} shows model effectiveness stratified by the number of rows in a table on partitions of WikiTableQuestions dev. We note that the 1024 token context window is often exceeded at 40-50 rows. Interestingly, we observe a universal degradation in performance well before 40 rows. As such, table size is an important factor in performance independent of model capacity.

For tables exceeding 50 rows model performance decreases by an average of 40\% relative to small tables. In these cases, due to prior ordering of the table, some questions only require information that is located at the top of the table. This maintains base performance for most models however we observe FlanT5-XL significantly degrades. Here we qualitatively observe hallucinations that repeat the input table tokens with large tables.

\subsection{Potential in row filtering}
\label{improvements_by_filtering}

Following our findings that table length has a significant and universal decrease in performance, we test the effect of filtering noise from the table. We hypothesize that removing noise from the data will increase model performance. We manually simulate a row-filtering tool by removing noise and only keeping the rows that are sufficient to answer the question. 

To test our hypothesis we manually annotate 51 samples of the WikiTableQuestions dev set. We validate that correct usage of a tool is critical by also simulating a random row filter that removes 50\% of rows indiscriminately. We note that the format for the input to the models is kept constant as we only change the number of rows in a table. Our findings are shown in Table \ref{tab:manual_subset} pertaining to correct and incorrect usage of a row filtering tool where we draw two conclusions: 

\textbf{Correct row-filtering markedly improves performance across all tested model types.} As tables reduce in size but retain relevant information, 4 of the 51 tables that are originally truncated to the 1024 token limit are fully processed. Noise is reduced by changing incorrect answers to positive ones  in 10\% of samples.

\textbf{Incorrect usage of a tool poses the risk of removing relevant information.} Since information in TQA is discreetly stored in table cells, it is challenging for models to recover once it is removed as we see by random row-filter performance. Furthermore, 7 of the 51 tables require no filtering. For such questions like \textit{“how many players participated?”} we observe that row-filtering is query-specific and demands detection strategies.

Our findings motivate an automatic appraoch to generate a query-specific row-filtering tool and to detect when to apply it.

\begin{table}[]
\centering
\resizebox{\columnwidth}{!}{%
\begin{tabular}{c|cccc}
\hline
\textbf{Row Filter} & \textbf{U.SKG} & \textbf{O.Tab} & \textbf{GPT-3} & \textbf{FlanT5} \\ \hline
\textbf{None}       & 39.21          & 54.9           & 47.05          & 17.64           \\ \hline
\textbf{Random}     & 11.76          & 5.88           & 15.64          & 5.52            \\ \hline
\textbf{Manual}     & \textbf{60.78} & \textbf{58.82} & \textbf{54.9}  & \textbf{39.21}  \\ \hline
\end{tabular}%
}
\caption{Exact match scores over 51 samples from the WikiTableQuestions dev set with gold rows selected manually as sufficient to answer the question.}
\label{tab:manual_subset}
\end{table}

\section{\shortname}

\shortname is our proposed method to address the limitations of current language models on large semi-structured data. In response to a question over a table, \shortname decides if tool-use is required. It then generates a program to transform and simplify the table to make question-answering more effective.  First, we outline our conceptual framework followed by our method implementation.

\subsection{Proposed model}

We introduce \shortname as $TW$ that combines a $Tool$ with a TQA model $F$ mediated by $\sigma \in [0,1]$ where $0$ means "don't apply tool" and $1$ means "apply tool".

\begin{align}
 TW(x, F) = \sigma \cdot & F(Tool(x)) + (1-\sigma) \cdot F(x) \label{tool_model}
\end{align}

Moreover, we define our $sigma$ as the output of a \textit{detector} function that aims to approximate the uncertainty of the model prediction.

\begin{align}
\sigma = d_\theta (x, F(x)) \approx P\left(y \neq F(x)\right)
\label{detector}
\end{align}

Our task is to maximize the exact match $EM$ over our corpus $D$.
We now take the partitions of $D$ over the correct and wrong predictions of the model $F$ (Eq. \ref{subsets}).

\begin{equation}
\begin{aligned}
\overline{S_F} = \{(x,\hat{y}) \in D | F(x) \neq \hat{y}\}  \\
S_F = \{(x,\hat{y}) \in D | F(x)=\hat{y}\} 
\label{subsets}
\end{aligned}
\end{equation}

As we intend, transforming an input $x$ with a tool might have a positive or negative effect on the produced output. When we observe our two subsets of $D$ we can draw the following conclusion.

\begin{align}
EM(TW(\cdot, F), \overline{S_F}) = \begin{cases}
0 & \sigma = 0 \\
\leq 1 & \sigma = 1
\end{cases}\label{incorrect_case} \\
EM(TW(\cdot, F), S_F) = \begin{cases}
1 & \sigma = 0 \\
\leq 1 & \sigma = 1
\end{cases} \label{correct_case}
\end{align}

The transformation on $x$ produced by the tool is guaranteed to increase or maintain performance when the model is known to be wrong (Eq \ref{incorrect_case}) and decrease it otherwise (Eq \ref{correct_case}). 
This justifies our choice in Eq \ref{detector} of a detector that approximates the probability for an incorrect prediction of model $F$.

\subsubsection{Tool use detection}
\label{detection_approach}

In an ideal setting, we maximize $EM$ over all available subsets: $S_F$ and $\overline{S_F}$. $\sigma$ mediates when we decide to use a tool.
For $S_F$ we can see the best choice is to use the original prediction yet for $\overline{S_F}$ we stand to gain if we use our tool before calling the model $F$. If we are in an oracle setting and we know the ground truth answer this gives us an oracle detector for when to use a tool.

\begin{align}
    \sigma = \mathbbm{1}(y \neq F(x))
\end{align}

However, in practice we approximate the oracle detector with a parametrized detector $d$. This aligns with previous work about query performance prediction \cite{cronen-townsend2002predictingqueryperformance} where estimating query difficulty is a reasonable assumption.

% \begin{align}
%     P(y \neq F(x)) \approx d_{\theta}(x, F(x))
% \end{align}

We explain our parametrized detector in more detail in Section \ref{methods}. In the extreme, however, $\sigma = 1$ equates to always applying the tool. Given that we are detecting when the model is likely to fail, how we define our tool directly impacts the performance we can have on the subset of $D$ it is applied over.

\subsubsection{Programs as tools}
\label{tool_approach}

The subset $\overline{S_F}$ is by definition difficult for $F$. As we see later in Section \ref{analysis}, our tested diverse set of state-of-the-art neural models (each corresponding to a specific instantiation of $F$) display similar patterns. This raises the natural following question: \textbf{What tool can distinctly complement the learned abilities of neural models?}

Our working hypothesis is that programs: 1) provide a natural interface to structured data; 2) circumvent several innate limitations of current neural systems due to their extrapolative nature. We define tools as short programs to transform the input data $x$ into $x^*$, which are in the same input domain $I$. The term $\epsilon$ is introduced to account for any noise introduced between the transformed input ($x^*$) and the original input domain ($I$). This is represented in Eq \ref{eq:tool_domain} as:

\begin{align}
    Tool &: I \rightarrow I^* + \epsilon \label{eq:tool_domain}
\end{align}

When paired selectively with an effective detection strategy $d$, programs as tools are applied to increase the likelihood of a correct prediction on a subset that is challenging for a model $F$. This outlines a program generation method $C(x)$ that generates a Python code dependent on the input. In Section \ref{methods} we describe our exploration into various methods for generating short programs as tools to interface in a query-specific way with structured data. 

\begin{align}
    Tool(x) = Exec[x, C(x)]
\end{align}

Deciding which programs to create as tools are strongly correlated with the target task as well as the limitations of the models. As such, we first clearly define our task in Section \ref{tasd_def} followed by an in-depth analysis of where our search for useful programs as tools will start.

\subsection{Model implementation}
\label{methods}

\shortname is composed of a tool use detector and a query-specific tool generator on top of an existing TQA model. Following our behavior analysis, OmniTab and UnifiedSKG act as our two best TQA models $F(x)$ for WikiTableQuestions and we use TapEx for WikiSQL.

\subsubsection{Model agnostic tool-use detector}
\label{sec:detector}

We develop a model-agnostic detector $d_{\theta}(x, F(x))$ to detect when a tool is likely to improve model accuracy. Detecting input difficulty is a reasonable assumption and aligns with previous work on quality estimation \cite{ueffing2005wordlevelconfidence, fomicheva2020unsupervisedquality} and query performance prediction \cite{cronen-townsend2002predictingqueryperformance}. The \textbf{combined} detector in \shortname is a linear classifier with the following features:

\smallskip
\textbf{Sequence log-probability (SeqLogProb)} is the length-normalised sequence log-probability from a trained model $F(y\mid y,x,\theta)$. 

\begin{align}
\frac{1}{L}\sum_{k=1}^{L}\log F(y_{k}\mid y_{<k},x,\theta)
\end{align}
\\
We expect low-confidence answers are likely to be incorrect.

\smallskip

\textbf{Input length}, as we have seen in Section \ref{analysis}, poses a challenge to all models irrespective of size and training objective. We leverage the size of the table measured by the number of rows as a simple feature to decide when to apply a tool. 

Our use of such simple detection methods contrasts with well-studied error detection methods in NLP \cite{berard2019naverlabseurope, fomicheva2020unsupervisedquality} as a sign that tools are reasonable model-agnostic extensions even with simple detection heuristics.

\subsubsection{Row-filter tool generator} 

The tool generator synthesizes a short Python program that takes a table as input and returns a transformed version of it. The following code snippet is fixed and highlights the area where the generated code from the tool generator is placed. 

\begin{figure}[!htb]
\centering
% \advance\lskip-4cm
\includegraphics[width=0.5\textwidth]{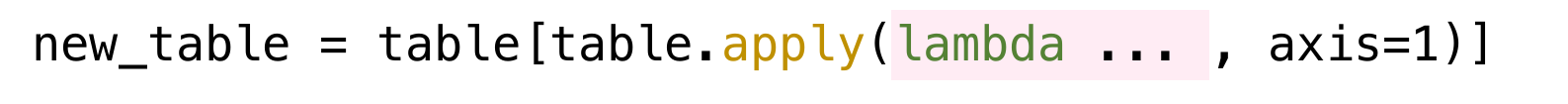}
\vspace{-3mm}
\label{fig:python-code}
\end{figure}

\begin{figure}%
    \centering
    \subfloat[\centering Zero-shot GPT-3 for the question "In how many games did sri lanka score at least 2 goals?"]{{\includegraphics[width=0.5\textwidth]{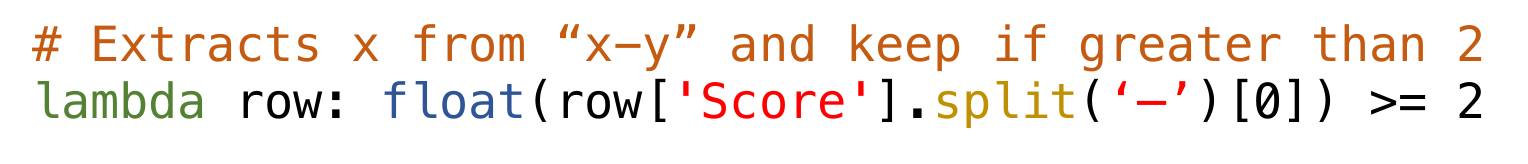} }}%
    \qquad
    \subfloat[\centering T5 for the question "Is France mentioned positively?"]{{\includegraphics[width=0.5\textwidth]{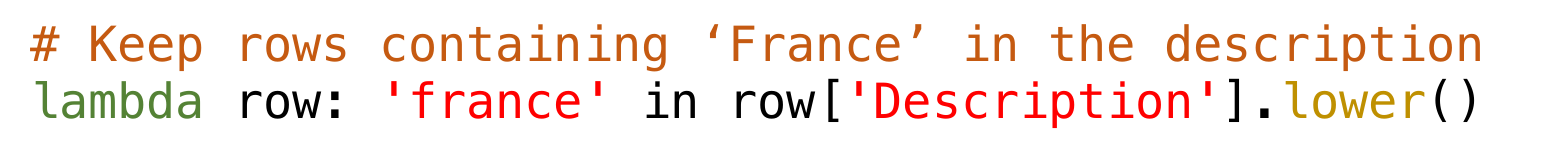} }}%
    \caption{Examples row filter tools generated from our two proposed methods. Comments are added manually for explanatory purposes.}%
    \label{fig:examples}%
\end{figure}

The task of the row filter generator is to generate a lambda function to remove the rows in the table that are not relevant to answering the question.
The program may be of arbitrary complexity emphasizing the generality of our approach for systems to interact with data through programs as seen in Figure \ref{fig:examples}. 

As we see in Table \ref{tab:manual_subset} removing rows requires care to preserve crucial information. It is evident that a tool must adapt its filtering strategy according to the question and the table. Our row filtering tool keeps its input and output space consistent and suitable for the downstream model.

The task of generating tools requires the model to produce an explicit transformation of the table given the question. Given that the search space for tools as programs grows exponentially with the expressiveness of the tools, we opt for methods that reduce the search space by having a prior on what possible transformations will work best. 

Specifically, we explore 2 approaches for generating Python row filters:

\smallskip
\textbf{Fine-tuned T5.} We fine-tune a T5 model through supervised training to autoregressively generate a python table filter given a question $q$ and a table $T$. For our supervised data we leverage a subset WikiTableQuestions \cite{pasupat2015compositionalsemantic} with SQUALL \cite{shi2020onthepotential} annotations on questions that contain a single \texttt{SELECT} and \texttt{WHERE} clause which are likely to benefit from row filtering.

\smallskip
\textbf{Zero-shot GPT-3.} We leverage GPT-3 for zero-shot prediction to generate a row filter as a Python lambda function.  We use the "text-davinci-003" API with a temperature of 0.2 with the question and table schema in the prompt (Appendix \ref{prompts}). Zero-shot tool generation shows the potential in low-effort approaches to manipulate structured data.

\medskip

Figure \ref{fig:examples} showcases generated Python code samples from both our proposed tool generators. Language models have no formal guarantees for executable code \cite{rae2021scalinglanguagemodels, chen2021evaluatinglargelanguage}. As a result, if the execution of the tool throws an exception or the resulting table is empty we revert to the original table. 

\smallskip
\textbf{Manual tool.} For WikiTableQuestions-Filter we leverage the SQUALL SQL annotations to derive a manual row filter. We analyze row filter headroom performance in Section \ref{simple_queries_results}.

\section{Results}

First, we investigate the various tool generators of \shortname on a subset of WikiTableQuestions where filtering is often required. Second, we focus on the importance of the detector to choose when to best apply the generated tools (Section \ref{when_to_apply}). Third, we test \shortname to both detect and generate tools across various TQA datasets and methods (Section \ref{main_results}). Finally, we analyze how \shortname performs as table size increases (Section \ref{length_analysis}).

\subsection{Performance of tool generators}
\label{simple_queries_results}

\begin{table}[]
\centering
\resizebox{\columnwidth}{!}{%
\begin{tabular}{cc|ll}
\hline
\multicolumn{1}{c|}{\textbf{Tool}}                      & \textbf{Detector} & \multicolumn{1}{c}{\textbf{Omnitab}} & \multicolumn{1}{c}{\textbf{UnifiedSKG}} \\ \hline \hline
\multicolumn{2}{c|}{\textbf{Baseline}}                                          & 73.5                                & 55.5                                   \\ \hline \hline
\multicolumn{1}{c|}{\multirow{2}{*}{\textbf{T5}}}       & \textbf{Always}   & 72.0 (-1.4)                         & 59.5 (+4.0)                            \\
\multicolumn{1}{c|}{}                                   & \textbf{Oracle}   & 77.5 (+4.0)                         & 63.3 (+7.8)                             \\ \hline
\multicolumn{1}{c|}{\multirow{2}{*}{\textbf{GPT-3}}} & \textbf{Always}   & 74.6 (+1.1)                          & 63.0 (+7.6)                            \\
\multicolumn{1}{c|}{}                                   & \textbf{Oracle}   & 80.3 (+6.8)                         & 68.6 (+13.1)                           \\ \hline
\multicolumn{1}{c|}{\multirow{2}{*}{\textbf{Human SQL}}}   & \textbf{Always}   & 74.9 (+1.4)                         & 65.0 (+9.5)                            \\
\multicolumn{1}{c|}{}                                   & \textbf{Oracle}   & 82.7 (+9.2)                         & 70.6 (+15.1)                           \\ \hline
\end{tabular}%
}
\caption{Row filter tool performance on WikiTableQuestions-Filtered with two detection strategies.}
\label{tab:simple_questions}
\end{table}

\begin{table*}[!h]
\centering
\begin{tabular}{lllllll}
\hline
                                                                                                   & \multicolumn{4}{c}{\textbf{WikiTableQuestions}}                                                             & \multicolumn{2}{c}{\textbf{WikiSQL}}    \\ \hline
\multicolumn{1}{l|}{}                                                                              & \multicolumn{2}{c|}{\textbf{OmniTab}}                & \multicolumn{2}{c|}{\textbf{UnifiedSKG}}             & \multicolumn{2}{c}{\textbf{TapEx}}      \\
\multicolumn{1}{l|}{\textbf{Detector}}                                                             & \multicolumn{1}{l|}{Dev} & \multicolumn{1}{l|}{Test} & \multicolumn{1}{l|}{Dev} & \multicolumn{1}{l|}{Test} & \multicolumn{1}{l|}{Dev} & Test         \\ \hline
\multicolumn{1}{l|}{\textbf{Never apply}}                                                          & 62.7                    & 63.0                     & 49.6                    & 50.8                     & 89.6                    & 89.0        \\
\multicolumn{1}{l|}{\textbf{Always apply}}                                                         & 56.5 (-6.2)             & 57.5 (-5.5)               & 48.5 (-1.0)             & 50.2 (-0.6)              & 89.6 (0.0)              & 89.8 (+0.7) \\
\multicolumn{1}{l|}{\textbf{SeqLogProb}}                                                           & 63.7 (+1.0)             & 64.3 (+1.2)              & 52.6 (+3.0)             & 54.6 (+3.8)               & 90.5 (+0.8)             & 90.4 (+1.3) \\
\multicolumn{1}{l|}{\textbf{Combined}}  & 63.7 (+1.0)             & 64.9 (+1.8)              & 52.9 (+3.4)             & 54.5 (+3.7)              & 90.7 (+1.1)             & 90.5 (+1.5) \\
\multicolumn{1}{l|}{\textbf{Oracle}}                                                              & 67.4 (+4.7)             & 68.3 (+5.3)              & 57.9 (+8.3)             & 59.0 (+8.1)              & 91.7 (+2.0)             & 91.5 (+2.4) \\ \hline
\end{tabular}
\caption{Exact match results on various detection strategies for applying our best row-filter tool generator: GPT-3.}
\label{tab:main_detector}
\end{table*}

As we observe in Table \ref{tab:simple_questions} our automatic tool generators (T5, zero-shot GPT-3) almost universally increase model performance on WikiTableQuestions-Filtered. Importantly, regardless of how our tools may be applied, UnifiedSKG significantly benefits by all tools generated by \shortname. This shows our tools are effective at filtering irrelevant information from tables that would otherwise cause TQA models to fail.

Manual tools show the potential for tool generators to simplify tables further. Our best tool generator, zero-shot GPT-3, achieves 70\% of manual performance averaged over all detection settings and models.

Table \ref{tab:simple_questions} also informs us of the importance of the detector. We observe a large gap for all tool generators comparing always applying our tool to oracle detection.

\subsection{Detecting when to use tools}
\label{when_to_apply}

Table \ref{tab:main_detector} shows performance of multiple detection strategies on the full dev and test sets for WikiTableQuestions and WikiSQL. We use our best-performing tool generator, zero-shot GPT-3. We observe row filtering tools require query specific detection since “always” or “never” applying tools shows the lowest results in all cases. 

We observe that even simple detection methods like SeqLogProb are sufficient to inform \shortname when to apply the query-specific generated row filter. We see significant benefits in leveraging tools for all TQA models in contrast to not using them. Performance increases further as we include table length as a feature in our detector highlighting the importance of using tools in accordance with the complexity of the data.

Under oracle detection conditions we observe significant potential for our generated tools. This shows how deciding \textbf{when} to apply a row filter tool is just as important as \textbf{how} to apply it.

\subsection{Overall performance}
\label{main_results}

Leveraging our prior findings, \shortname is the combination of our best detection method (SeqLogProb with table length) and our best row-filter tool generator (zero-shot GPT-3). For each dataset, we show the corresponding model $F$ as our base TQA model. Table \ref{tab:main_WTQ} and Table \ref{tab:main_WSQL} show overall model performance on WikiTableQuestions and WikiSQL respectively. 

Our results show \shortname significantly improves performance agnostic of the target model using the generated tools. UnifiedSKG is particularly effective in leveraging the transformed tables with a 3.6\% absolute performance increase compared to not using tools. When paired with OmniTab and TapEx we improve the state-of-the-art for both datasets. The improvement in WikiSQL is particularly impactful as \shortname enables a 10\% error-rate reduction.

These results show how programmatic tools effectively complement neural components as an effective method for processing semi-structured data. In the following section, we perform a stratified analysis to understand where \shortname leads to the most improvements.

\begin{table}[]
\centering
\resizebox{\columnwidth}{!}{%
\begin{tabular}{l|l|l}
\hline
\textbf{Method}     & \textbf{Dev}                                            & \textbf{Test}                                           \\ \hline
2-shot GPT-3 Direct \cite{Chen2022LargeLM} & ---                                                     & 27.3                                                    \\ \hline
BART \cite{lewis2020bartdenoisingsequence}                & 37.2                                                    & 38.0                                                    \\ \hline
2-shot GPT-3 CoT  \cite{Chen2022LargeLM}  & ---                                                     & 45.7                                                    \\ \hline
UnifiedSKG \cite{Xie2022UnifiedSKGUA}          & \begin{tabular}[c]{@{}l@{}}50.9 \\ (49.6)\end{tabular} & \begin{tabular}[c]{@{}l@{}}50.9 \\ (50.8)\end{tabular} \\ \hline
\textbf{ToolWriter} + UnifiedSKG              & 52.9                                                    & 54.5                                                    \\ \hline
TapEx \cite{Liu2021TAPEXTP}              & 57.0                                                    & 57.5                                                    \\ \hline
OmniTab  \cite{Jiang2022OmniTabPW}           & \begin{tabular}[c]{@{}l@{}}--- \\ (62.7)\end{tabular}  & \begin{tabular}[c]{@{}l@{}}62.8 \\ (63.0)\end{tabular} \\ \hline
\textbf{ToolWriter } + Omnitab & \textbf{63.7}                                          & \textbf{64.9}                                          \\ \hline
\end{tabular}%
}
\caption{Exact match accuracy results on WikiTableQuestions. Results in parenthesis are our reproduced experiments.}
\label{tab:main_WTQ}
\end{table}

\begin{table}[]
\centering
\resizebox{\columnwidth}{!}{%
\begin{tabular}{l|l|l}
\hline
\textbf{Method}     & \textbf{Dev}                                            & \textbf{Test}                                           \\ \hline
BART     \cite{lewis2020bartdenoisingsequence}           & 87.3                                                    & 85.8                                                    \\ \hline
UnifiedSKG  \cite{Xie2022UnifiedSKGUA}        & 87.4                                                   & 85.7                                                   \\ \hline
OmniTab  \cite{Jiang2022OmniTabPW}           & ---                                                     & 88.7                                                    \\ \hline
TapEx   \cite{Liu2021TAPEXTP}            & \begin{tabular}[c]{@{}l@{}}89.2 \\ (89.6)\end{tabular} & \begin{tabular}[c]{@{}l@{}}89.5 \\ (89.0)\end{tabular} \\ \hline
\textbf{ToolWriter} + TapEx & \textbf{90.7}                                          & \textbf{90.5}                                          \\ \hline
\end{tabular}%
}
\caption{Exact match accuracy results on WikiSQL. Results in parenthesis are our reproduced experiments. }
\label{tab:main_WSQL}
\end{table}

\subsection{Tools improve performance on long tables}
\label{length_analysis}

\begin{table*}[]
\centering
\begin{tabular}{llllll}
\hline
\textbf{}                                                   & \textbf{}                                & \multicolumn{2}{c|}{\textbf{Omnitab}}                & \multicolumn{2}{c}{\textbf{UnifiedSKG}}         \\ \hline
                                                            &                                          & \multicolumn{1}{l|}{Dev} & \multicolumn{1}{l|}{Test} & \multicolumn{1}{l|}{Dev} & Test                 \\ \hline
\multicolumn{1}{l|}{\multirow{2}{*}{rows $<$ 30}}           & \multicolumn{1}{l|}{\textbf{Baseline}}   & 67.0                     & 67.9                      & 52.7                     & 54.3                 \\
\multicolumn{1}{l|}{}                                       & \multicolumn{1}{l|}{\textbf{ToolWriter}} & 67.0 (+0.0)              & \textbf{68.0 (+0.1)}      & \textbf{55.0 (+2.3)}     & \textbf{57.2 (+2.8)} \\ \hline
\multicolumn{6}{l}{}                                                                                                                                                                                            \\ \hline
\multicolumn{1}{l|}{\multirow{2}{*}{30 $\leq$ rows $<$ 60}} & \multicolumn{1}{l|}{\textbf{Baseline}}   & 48.7                     & 46.4                      & 39.6                     & 37.5                 \\
\multicolumn{1}{l|}{}                                       & \multicolumn{1}{l|}{\textbf{ToolWriter}} & \textbf{52.6 (+3.9)}     & \textbf{51.9 (+5.5)}      & \textbf{44.0 (+4.4)}     & \textbf{45.1 (+7.6)} \\ \hline
\multicolumn{6}{l}{}                                                                                                                                                                                            \\ \hline
\multicolumn{1}{l|}{\multirow{2}{*}{rows $\leq$ 60}}        & \multicolumn{1}{l|}{\textbf{Baseline}}   & 41.4                     & 35.0                      & 33.7                     & 34.3                 \\
\multicolumn{1}{l|}{}                                       & \multicolumn{1}{l|}{\textbf{ToolWriter}} & \textbf{48.1 (+6.6)}     & \textbf{43.3 (+8.3)}      & \textbf{42.0 (+8.3)}     & \textbf{41.7 (+7.5)} \\ \hline
\end{tabular}%

\caption{Row filtering performance comparison on partitions stratified by table length for WikiTableQuestions.}
\label{tab:row_analysis}
\end{table*}

In this section we do an ablation study stratified by table length on WikiTableQuestions: short tables (rows $<$ 30), medium tables (30 $\leq$ rows $<$ 60), and long tables (60 $\leq$ rows). We aim to quantify the effect \shortname has as table size increases. As in Section \ref{main_results} \shortname uses GPT-3 as the tool generator and the combined detector.

Table \ref{tab:row_analysis} shows our original hypothesis confirmed: Row filtering tools can be an effective strategy to help models handle long tables. 
We notice how as table length increases, the positive effect of the row filtering tool becomes more pronounced. Our hypothesis is further confirmed with our T5 tool generator where results mimic the Table \ref{tab:row_analysis} reaching up to 5\% absolute improvement with UnifiedSKG. 

As noted in Section \ref{sec:detector}, detection is critical to tool-use. On short tables, we observe no degradation in performance highlighting the effectiveness of our combined detector.

\section{Background and Related Work}

Semantic parsing focuses on generating an executable parse for the exact answer \cite{McClelland1986MechanismsOS}, benefiting from data size independence \cite{Herzig2017NeuralSP}. It requires strong supervision \cite{Dong2018CoarsetoFineDF, Yin2021CompositionalGF} or reinforcement learning \cite{zhong2017seq2sqlgenerating} and assumes coherent data formatting and an expressive target language.

Alternative approaches learn a joint table-question-answer mapping. Seq2Seq models \cite{Sutskever2014SequenceTS} execute \cite{Zaremba2014LearningTE} and simulate formal programs \cite{Lu2015NeuralEL}. Intermediate executable modules were integrated \cite{Neelakantan2015NeuralPI}, while Transformer-based models \cite{vaswani2017attention, lewis2020bartdenoisingsequence, raffel2020exploringthelimits} leveraged unsupervised language capabilities \cite{xie2022unifiedskgunifying, Jiang2022OmniTabPW, Herzig2020TaPasWS, Yin2020TaBERTPF}.

Recent interest in execution-loop models arises from language models' ability to explain reasoning \cite{wei2022chainofthought}, improving compositional questions \cite{Zhou2022LeasttoMostPE} and symbolic manipulation \cite{Bueno2022InducedNL, nye2021showyourwork, Wolfson2020BreakID}. TQA language models generate chains of thought with sub-question answers \cite{Chen2022LargeLM}.

Recent advances in code-focused language models led to an interest in combining question decomposition and program interaction \cite{chen2021evaluatinglargelanguage}. Toolformer \cite{Schick2023ToolformerLM}, Program Assisted Learning \cite{Gao2022PALPL}, and Tool Augmented Language Models \cite{parisi2022talmtoolaugmented} interleave execution and natural language reasoning but face limitations in capacity. Our work addresses large structured context directly while interleaving execution and natural language.

\section{Conclusion}

Tabular question answering is a challenging setting for neural methods due to large context sizes and implicit reasoning. First, we characterize the limitations of neural methods to integrate structured data and find all language modeling methods struggle with large tables. Second, we propose \shortname to generate query-specific tools to simplify large tables and detect when these transformations should be applied. We propose various language model-based methods to generate programs that filter rows which universally improve and achieve state-of-the-art results on two tabular question-answering datasets. Finally, we determine significant headroom in both detecting when to use tools and how to generate them under oracle setting highlighting the potential in tools to manipulate structured data combined with language models.

\bibliography{main, custom, semantic_scholar}
\bibliographystyle{acl_natbib}

\appendix

\section{Experimental setting}
\label{experimental_setting}

We use the recommended settings for each and linearize the tables with the official scripts. At test time we perform greedy decoding. We obtain these models from the Huggingface model hub \cite{wolf2019huggingfacestransformers}. 

Our two tool generation models: prompt-based and fine-tuned. For our prompt-based model we use the OpenAI GPT-3 API with the "text-davinci-003" model with a temperature of 0.2. 

\section{Subset analysis}

To effectively filter according to relevant criteria we leverage parallel SQL annotations from SQUALL \cite{shi2020onthepotential} which cover 77.11\% of the dev data. These annotations are formal semantic parses of the query which remove the natural language variability enabling us to filter by required capability. 

As shown in the first two rows of Table \ref{tab:wtq-analysis} we see a marked decrease in performance for non SQUALL annotated samples across all model types. These are cases where the queries or table are considered too complex to be expressed as SQL. As such we are restricted to qualitative and point-wise analysis of these samples to characterize model behavior.

\begin{table*}
\centering
\begin{tabular}{llllll}
\hline
\textbf{Data Subset}      & \textbf{Uni.SKG}              & \textbf{OmniTab}              & \textbf{GPT-3}                & \textbf{FlanT5}               & \textbf{dataset \%} \\ \hline
\textbf{SQUALL annotated} & \cellcolor[HTML]{FFFFFF}54.56 & \cellcolor[HTML]{FFFFFF}66.65 & \cellcolor[HTML]{FFFFFF}38.62 & \cellcolor[HTML]{FFFFFF}17.36 & 77.11               \\ \hline
Non SQUALL annotated      & \cellcolor[HTML]{F5CCC8}33.49 & \cellcolor[HTML]{F5CBC7}40.43 & \cellcolor[HTML]{F9E3E1}30.4  & \cellcolor[HTML]{F6D1CD}11.27 & 22.89               \\ \hline
+ 1' as offset            & \cellcolor[HTML]{57BB8A}69.23 & \cellcolor[HTML]{57BB8A}73.85 & \cellcolor[HTML]{A4DAC0}47.69 & \cellcolor[HTML]{FEFBFB}16.92 & 2.3                 \\ \hline
requires counting rows    & \cellcolor[HTML]{F9E3E1}43.14 & \cellcolor[HTML]{FEFDFD}66.11 & \cellcolor[HTML]{F7D8D5}27.17 & \cellcolor[HTML]{EFAEA9}6.72  & 25.22               \\ \hline
count all rows            & \cellcolor[HTML]{F8DBD8}39.62 & \cellcolor[HTML]{FFFFFF}66.67 & \cellcolor[HTML]{F0B5B0}16.98 & \cellcolor[HTML]{EB9891}3.77  & 5.62                \\ \hline
big sub or add            & \cellcolor[HTML]{E98B83}6.58  & \cellcolor[HTML]{EA938B}11.84 & \cellcolor[HTML]{57BB8A}55.26 & \cellcolor[HTML]{E98F88}2.63  & 2.68                \\ \hline
1 'where' clause          & \cellcolor[HTML]{F3FAF7}55.65 & \cellcolor[HTML]{A0D9BD}70.75 & \cellcolor[HTML]{F2FAF6}39.92 & \cellcolor[HTML]{A8DCC3}17.63 & 44.68               \\ \hline
2 'where' clauses         & \cellcolor[HTML]{FCF2F1}49.42 & \cellcolor[HTML]{FCF1F0}59.83 & \cellcolor[HTML]{E1F3EA}41.62 & \cellcolor[HTML]{FCEFEE}15.32 & 12.22               \\ \hline
count and where           & \cellcolor[HTML]{FEFEFE}54.41 & \cellcolor[HTML]{E5F5ED}67.8  & \cellcolor[HTML]{F4FBF8}39.73 & \cellcolor[HTML]{FDF7F6}16.35 & 63.3                \\ \hline
answer not in prompt      & \cellcolor[HTML]{F3C4C0}30.23 & \cellcolor[HTML]{F1B9B4}31.4  & \cellcolor[HTML]{F7D6D3}26.74 & \cellcolor[HTML]{EB9992}3.88  & 9.11                \\ \hline
multiple answers          & \cellcolor[HTML]{F9E0DE}41.94 & \cellcolor[HTML]{F9E1DF}51.61 & \cellcolor[HTML]{FAE9E7}32.26 & \cellcolor[HTML]{E67C73}0     & 2.19                \\ \hline
one select                & \cellcolor[HTML]{F0F9F5}55.92 & \cellcolor[HTML]{D8F0E4}68.33 & \cellcolor[HTML]{FEFDFD}38.15 & \cellcolor[HTML]{57BB8A}17.88 & 63.79               \\ \hline
one select and one where  & \cellcolor[HTML]{EFF9F4}55.99 & \cellcolor[HTML]{95D5B6}71.2  & \cellcolor[HTML]{F1FAF5}40.06 & \cellcolor[HTML]{92D3B3}17.7  & 43.91               \\ \hline
has duplicate columns     & \cellcolor[HTML]{F4C5C1}30.59 & \cellcolor[HTML]{F5CCC9}41.18 & \cellcolor[HTML]{F3C3BF}21.18 & \cellcolor[HTML]{F5CBC8}10.59 & 3                   \\ \hline
\end{tabular}
\caption{Exact match performance of systems according to subsets of the WikiTableQuestions dev set.}
\label{tab:wtq-analysis}
\end{table*}

\section{Prompts}
\label{prompts}

\begin{figure*}
    \centering
\begin{lstlisting}
User 1:
I need an expert to help me answer the question by making the table 
smaller.
Question: Who are all of the players on the Westchester High School 
club team?

table = {'Player': ['Jarrett Jack', 'Jermaine Jackson', ...
'No.': ['1', '8', ...
'Nationality': ['United States', 'United States', ...
'Position': ['Guard', 'Guard', ...
'Years in Toronto': ['2009-10', '2002-03', ...
'School/Club Team': ['Georgia Tech', 'Detroit', ...
}

User 2:
For 'Who are all of the players on the Westchester High School club 
team?' the most impactful change will be to filter the rows. Since I 
don't know all the rows I'll use rough string matching, float casting, 
lowering and be as broad as possible.

>>> new_table = table[table.apply(lambda row_dict:|\colorbox{magenta!30}{ }|'Westchester' in 
row_dict['School/Club Team'].lower(), axis=1)]
\end{lstlisting}
\caption{Prompt used to generate row filter tools with GPT-3 in a zero-shot setup. Tables are truncated to 2 rows to give the model a schema for how to interact with the data. Hilighted region indicates the start of the prompt completion.}
    \label{fig:my_label}
\end{figure*}

\end{document}